\documentclass[11pt,a4paper]{article}
\usepackage[hyperref]{acl2021}
\usepackage{times}
\usepackage{latexsym}
\usepackage{graphicx}
\usepackage{url}
\usepackage{booktabs}

\usepackage{stfloats}

\usepackage{amsfonts,amssymb}
\usepackage{mydef}
\usepackage{xcolor}
\usepackage{tabularx}
\usepackage[figuresright]{rotating}
\usepackage{multirow}

\usepackage{microtype}
\aclfinalcopy
\title{Paradigm Shift in Natural Language Processing}

\author{Tianxiang Sun, Xiangyang Liu, Xipeng Qiu, Xuanjing Huang \\
  School of Computer Science, Fudan University \\
  Shanghai Key Laboratory of Intelligent Information Processing, Fudan University \\
  \texttt{\{txsun19,xiangyangliu20,xpqiu,xjhuang\}@fudan.edu.cn} \\}

\begin{document}

\maketitle


\begin{abstract}
In the era of deep learning, modeling for most NLP tasks have converged to several mainstream paradigms. For example, we usually adopt the sequence labeling paradigm to solve a bundle of tasks such as POS-tagging, NER, Chunking, and adopt the classification paradigm to solve tasks like sentiment analysis. With the rapid progress of pre-trained language models, recent years have observed a rising trend of \textit{Paradigm Shift}, which is solving one NLP task by reformulating it as another one. Paradigm shift has achieved great success on many tasks, becoming a promising way to improve model performance. Moreover, some of these paradigms have shown great potential to unify a large number of NLP tasks, making it possible to build a single model to handle diverse tasks. In this paper, we review such phenomenon of paradigm shifts in recent years, highlighting several paradigms that have the potential to solve different NLP tasks.\footnote{A constantly updated website is publicly available at \href{https://txsun1997.github.io/nlp-paradigm-shift}{https://txsun1997.github.io/nlp-paradigm-shift}}

% Traditionally, each NLP task usually has its own processing paradigm. For example, we usually use sequence labeling paradigm to solve the NER task and use classification paradigm to solve the sentiment analysis task. With the rapid progress of deep learning and pretrained model, many studies found we can solve a NLP task in different paradigms. In this talk, we summarize the current NLP paradigms into six categories: classification, matching, sequence labeling, span extraction, generation, action generation. Then we discuss the paradigm shift phenomenon among various NLP tasks.
\end{abstract}

\section{Introduction}\label{sec:intro}
% Solving a NLP task is typically to build a machine learning model to fit the data set $\{x_i,y_i\}_{i=1}^N$ corresponding to the task, where $x_i$ can be structured text and $y_i$ can be structured label. The structure of the text and label reflects the nature of the task, hence is usually different across tasks.
\textit{Paradigm} is the general framework to model a class of tasks. For instance, sequence labeling is a mainstream paradigm for named entity recognition (NER). Different paradigms usually require different input and output, therefore highly depend on the annotation of the tasks. In the past years, modeling for most NLP tasks have converged to several mainstream paradigms, as summarized in this paper, \texttt{Class}, \texttt{Matching}, \texttt{SeqLab}, \texttt{MRC}, \texttt{Seq2Seq}, \texttt{Seq2ASeq}, and \texttt{(M)LM}.

Though the paradigm for many tasks has converged and dominated for a long time, recent work has shown that models under some paradigms also generalize well on tasks with other paradigms. For example, the \texttt{MRC} paradigm and the \texttt{Seq2Seq} paradigm can also achieve state-of-the-art performance on NER tasks~\cite{Li20MRC,Yan2021UniNER}, which are previously formalized in the sequence labeling (\texttt{SeqLab}) paradigm. Such methods typically first convert the form of the dataset to the form required by the new paradigm, and then use the model under the new paradigm to solve the task. In recent years, similar methods that reformulate a NLP task as another one have achieved great success and gained increasing attention in the community. After the emergence of the pre-trained language models (PTMs) ~\cite{devlin-etal-2019-bert,Raffel20T5,Brown2020GPT3,qiu2020:scts-ptms}, paradigm shift has been observed in an increasing number of tasks. Combined with the power of these PTMs, some paradigms have shown great potential to unify diverse NLP tasks. One of these potential unified paradigms, \texttt{(M)LM} (also referred to as \textit{prompt-based tuning}), has made rapid progress recently, making it possible to employ a single PTM as the universal solver for various understanding and generation tasks~\cite{Schick21PET,Schick21Size,Gao20Making,Shin20Autoprompt,Li20Prefix,Liu21PTuning,Lester21Power}.

Despite their success, these paradigm shifts scattering in various NLP tasks have not been systematically reviewed and analyzed. In this paper, we attempt to summarize recent advances and trends on this line of research, namely \textit{paradigm shift} or \textit{paradigm transfer}.

This paper is organized as follows. In section~\ref{sec:paradigms}, we give formal definitions of the seven paradigms, and introduce their representative tasks and instance models. In section~\ref{sec:paradigm_shift}, we show recent paradigm shifts happened in different NLP tasks. In section~\ref{sec:unified_paradigm}, we discuss designs and challenges of several highlighted paradigms that have great potential to unify most existing NLP tasks. In section~\ref{sec:conclusion}, we conclude with a brief discussion of recent trends and future directions.

% We call this phenomena \textit{paradigm shift} or \textit{paradigm transfer}. In recent years, paradigm shifts such as \texttt{SeqLab}~$\rightarrow$~\texttt{MRC} have achieved great success and gained increasing attention in the community. In this survey, we attempt to summarize recent advances of paradigm shift in NLP.

% The modeling for each category of tasks with the same structure of input and output has converged as a mainstream \textit{paradigm}. The structures of data sets and their corresponding modeling paradigms are shown in Table 1.

% For a long time, the paradigm for each category of tasks, \textit{e.g.} sequence labeling for NER and POS-Tagging, is dominant in these tasks. However, recent years have also witnessed great power of the paradigm transfer, which is to frame a task as another task by changing the form of the problem. In this paper, we summarize these successful paradigm shift work and shed some light on the potential use of paradigm transfer.

\section{Paradigms in NLP}
\label{sec:paradigms}
% In this section, we will give formal definitions of several mainstream paradigms, and their corresponding tasks and representative models. In Sec.~\ref{sec:ParaTaskModel}, we make clear the relationship among paradigms, tasks, and models; In Sec.~\ref{sec:categories}, we summarize seven categories of paradigms, and give their formal definitions, representative tasks, and instance models. In Sec.~\ref{sec:compound}, we show how complex NLP tasks can be modeled by combining multiple paradigms.

\subsection{Paradigms, Tasks, and Models}
\label{sec:ParaTaskModel}
Typically, a task corresponds to a dataset $\mathcal{D}=\{\cX_i,\cY_i\}_{i=1}^N$. Paradigm is the general modeling framework to fit some datasets (or tasks) with a specific format (\textit{i.e.}, the data structure of $\cX$ and $\cY$). Therefore, a task can be solved by multiple paradigms by transforming it into different formats, and a paradigm can be used to solve multiple tasks that can be formulated as the same format. A paradigm can be instantiated by a class of models with similar architectures.
% \subsection{Notations}
% \label{sec:notation}
% \begin{itemize}
%   \item $x$ denotes a symbol;
%   \item $\cX={x_{1},\cdots,x_{L}}$ denotes a sequence with length $L$;
%   \item $\bx$ denotes a vector indicating the embedding of symbol $x$;
%   \item $\cY$ denotes the label for an input sequence;
%   \item $f_{\theta}$ denotes a model parameterized by $\theta$;
% \end{itemize}

\subsection{The Seven Paradigms in NLP}
\label{sec:categories}
\begin{figure*}[t!]
    \centering
    \includegraphics[width=.8\linewidth]{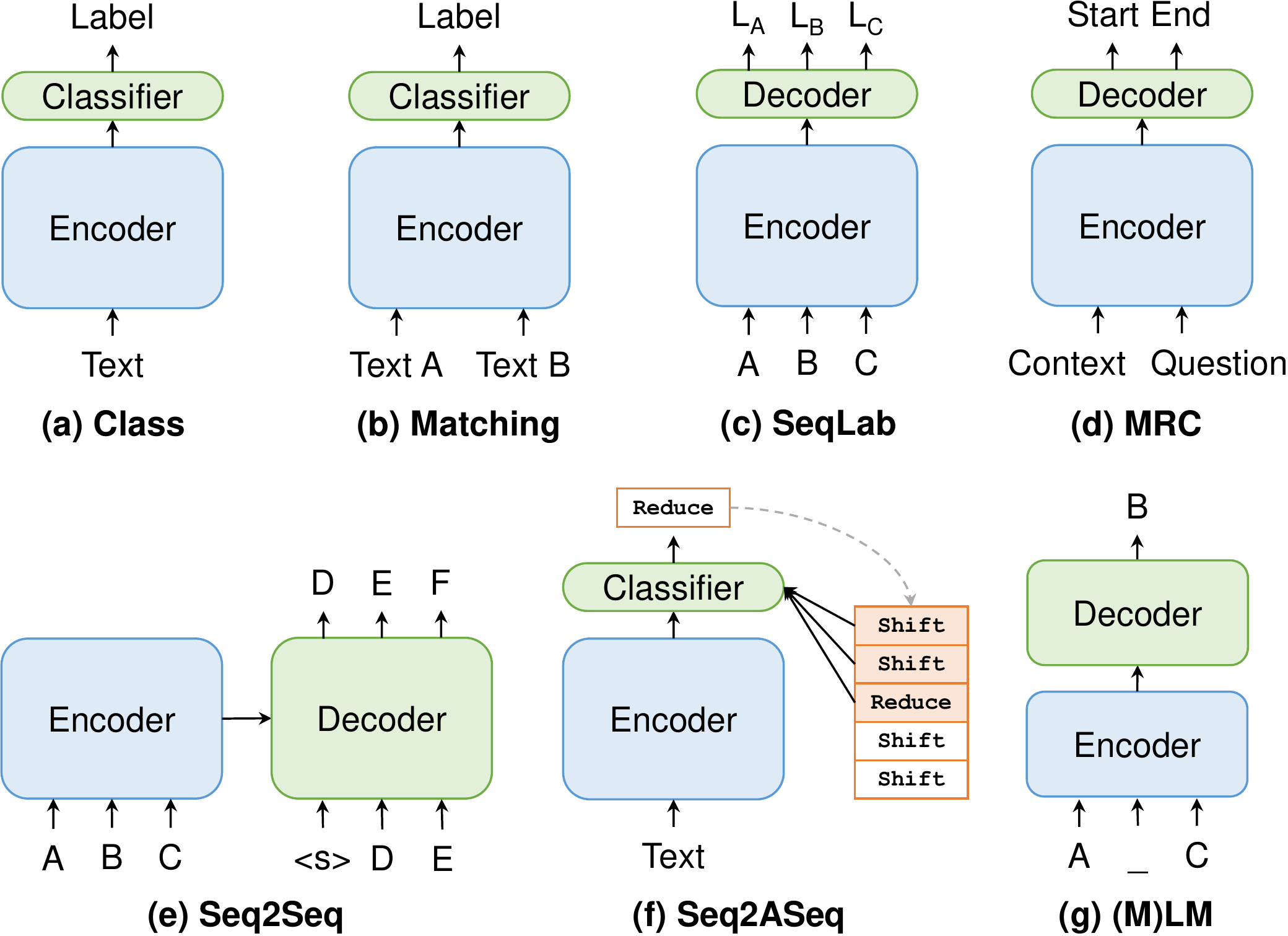}
    \caption{Illustration of the seven mainstream paradigms in NLP.}
    \label{fig:paradigms}
\end{figure*}

In this paper, we mainly consider the following seven paradigms that are widely used in NLP tasks, i.e. \texttt{Class}, \texttt{Matching}, \texttt{SeqLab}, \texttt{MRC}, \texttt{Seq2ASeq}, and \texttt{(M)LM}. These paradigms have demonstrated strong dominance in many mainstream NLP tasks. In the following sections, we briefly introduce the seven paradigms and their corresponding tasks and models.

\subsubsection{Classification (\texttt{Class})}
Text classification, which is designating predefined labels for text, is an essential and fundamental task in various NLP applications such as sentiment analysis, topic classification, spam detection, \textit{etc}. In the era of deep learning, text classification is usually done by feeding the input text into a deep neural-based encoder to extract the task-specific feature, which is then fed into a shallow classifier to predict the label, \textit{i.e.}
\begin{equation}
    \cY = \textsc{Cls}(\textsc{Enc}(\cX)).
\end{equation}
Note that $\cY$ can be one-hot or multi-hot (in which case we call multi-label classification). $\textsc{Enc}(\cdot)$ can be instantiated as convolutional networks~\cite{Kim2014Conv}, recurrent networks~\cite{Liu2016Recurrent}, or Transformers~\cite{Vaswani2017Attention}. $\textsc{Cls}(\cdot)$ is usually implemented as a simple multi-layer perceptron following a pooling layer. Note that the pooling layer can be performed on the whole input text or a span of tokens.

\subsubsection{\texttt{Matching}}
Text matching is a paradigm to predict the semantic relevance of two texts. It is widely adopted in many fields such as information retrieval, natural language inference, question answering and dialogue systems. A matching model should not only extract the features of the two texts, but also capture their fine-grained interactions. The \texttt{Matching} paradigm can be simply formulated as
\begin{equation}
    \cY = \textsc{Cls}(\textsc{Enc}(\cX_a, \cX_b)),
\end{equation}
where $\cX_a$ and $\cX_b$ are two texts to be predicted, $\cY$ can be discrete (\textit{e.g.} whether one text entails or contradicts the other text) or continuous (\textit{e.g.} semantic similarity between the two texts). The two texts can be separately encoded and then interact with each other~\cite{Chen2017ESIM}, or be concatenated to be fed into a single deep encoder~\cite{devlin-etal-2019-bert}.

\subsubsection{Sequence Labeling (\texttt{SeqLab})}

The Sequence Labeling (\texttt{SeqLab}) paradigm (also referred to as Sequence Tagging) is a fundamental paradigm modeling  a variety of tasks such as part-of-speech (POS) tagging, named entity recognition (NER), and text chunking. Conventional neural-based sequence labeling models are comprised of an encoder to capture the contextualized feature for each token in the sequence, and a decoder to take in the features and predict the labels, \textit{i.e.}
\begin{equation}
    y_1, \cdots, y_n = \textsc{Dec}(\textsc{Enc}(x_1, \cdots, x_n)),
\end{equation}
where $y_1, \cdots, y_n$ are the corresponding labels of $x_1, \cdots, x_n$. $\textsc{Enc}(\cdot)$ can be instantiated as a recurrent network~\cite{Ma16End2end} or a Transformer encoder~\cite{Vaswani2017Attention}. $\textsc{Dec}(\cdot)$ is usually implemented as conditional random fields (CRF)~\cite{Lafferty2001CRF}.

\subsubsection{\texttt{MRC}}

Machine Reading Comprehension (\texttt{MRC}) paradigm extracts contiguous token sequences (spans) from the input sequence conditioned on a given question. It is initially adopted to solve MRC task, then is generalized to other NLP tasks by reformulating them into the MRC format. Though, to keep consistent with prior work and avoid confusion, we name this paradigm \texttt{MRC}, and distinguish it from the task MRC. The \texttt{MRC} paradigm can be formally described as follows,
\begin{equation}
    y_k\cdots y_{k+l} = \textsc{Dec}(\textsc{Enc}(\cX_{p}, \cX_{q}))
\end{equation}
where $\cX_{p}$ and $\cX_{q}$ denote passage (also referred to context) and query, and $y_k\cdots y_{k+l}$ is a span from $\cX_{p}$ or $\cX_{q}$. Typically, $\textsc{Dec}$ is implemented as two classifiers, one for predicting the starting position and one for predicting the ending position~\cite{Xiong2017DCN,Seo2017Bidaf,Chen2017DrQA}.

\subsubsection{Sequence-to-Sequence (\texttt{Seq2Seq})}

Sequence-to-Sequence (\texttt{Seq2Seq}) paradigm is a general and powerful paradigm that can handle a variety of NLP tasks. Typical applications of \texttt{Seq2Seq} include machine translation and dialogue, where the system is supposed to output a sequence (target language or response) conditioned on a input sequence (source language or user query). \texttt{Seq2Seq} paradigm is typically implemented by an encoder-decoder framework~\cite{Sutskever2014Seq2seq,Bahdanau2015NMT,Luong2015Multitask,Gehring2017Convseq2seq}:
\begin{equation}
    y_1, \cdots, y_m = \textsc{Dec}(\textsc{Enc}(x_1, \cdots, x_n)).
\end{equation}
% Different to sequence labeling, the predicted sequence $\cY$ does not necessarily be aligned to $\cX$ in Seq2Seq paradigm. Typical NLP tasks in seq2seq paradigm include machine translation and dialogue systems.
Different from \texttt{SeqLab}, the lengths of the input and output are not necessarily the same. Moreover, the decoder in \texttt{Seq2Seq} is usually more complicated and takes as input at each step the previous output (when testing) or the ground truth (with teacher forcing when training).

\subsubsection{Sequence-to-Action-Sequence (\texttt{Seq2ASeq})}
Sequence-to-Action-Sequence (\texttt{Seq2ASeq}) is a widely used paradigm for structured prediction. The aim of \texttt{Seq2ASeq} is to predict an action sequence (also called transition sequence) from some initial configuration $c_0$ to a terminal configuration. The predicted action sequence should encode some legal structure such as dependency tree. The instances of the \texttt{Seq2ASeq} paradigm are usually called transition-based models, which can be formulated as
\begin{equation}
    {\cA} = \textsc{Cls}(\textsc{Enc}(\cX), \cC),
\end{equation}
where $\cA = a_1, \cdots, a_m$ is a sequence of actions, $\cC = c_0, \cdots, c_{m-1}$ is a sequence of configurations. At each time step, the model predicts an action $a_t$ based on the input text and current configuration $c_{t-1}$, which can be comprised of top elements in stack, buffer, and previous actions~\cite{Chen2014Fast,Dyer2015Transition}.

% \begin{align}
% \hat{\cY} &= \argmax_{\cY \in \mathrm{gen}(\cX)} \mathrm{score}(\cX,\cY)\\
% &= \argmax_{a_0,\cdots,a_m\rightarrow \cY} \sum_{i=0}^{m} \mathrm{score}(\cX, h_i, a_i),
% \end{align}
% where $\mathrm{gen}(\cX)$ denotes the set of all possible structure from $\cX$, $a_i$ denotes the action, and $h_i$ denotes the state at step $i$.

% stack

% tree-lstm

\subsubsection{\texttt{(M)LM}}
Language Modeling (LM) is a long-standing task in NLP, which is to estimate the probability of a given sequence of words occurring in a sentence. Due to its self-supervised fashion, language modeling and its variants, \textit{e.g.} masked language modeling (MLM), are adopted as training objectives to pre-train models on large-scale unlabeled corpus. Typically, a language model can be simply formulated as
\begin{equation}
    x_k = \textsc{Dec}(x_1, \cdots, x_{k-1}),
\end{equation}
where $\textsc{Dec}$ can be any auto-regressive model such as recurrent networks~\cite{Bengio2000NeuralLM,Grave2017Improve} and Transformer decoder~\cite{Dai2019TransformerXL}. As a famous variant of LM, MLM can be formulated as
\begin{equation}
    \bar{x} = \textsc{Dec}(\textsc{Enc}(\tilde{x})),
\end{equation}
where $\tilde{x}$ is a corrupted version of $x$ by replacing a portion of tokens with a special token \texttt{[MASK]}, and $\bar{x}$ denotes the masked tokens to be predicted. $\textsc{Dec}$ can be implemented as a simple classifier as in BERT~\cite{devlin-etal-2019-bert} or an auto-regressive Transformer decoder as in BART~\cite{bart} and T5~\cite{Raffel20T5}.

Though LM and MLM can be somehow different (LM is based on auto-regressive while MLM is based on auto-encoding), we categorize them into one paradigm, \texttt{(M)LM}, due to their same inherent nature, which is estimating the probability of some words given the context.

\subsection{Compound Paradigm}
\label{sec:compound}
In this paper, we mainly focus on fundamental paradigms (as described above) and tasks. Nevertheless, it is worth noting that more complicated NLP tasks can be solved by combining multiple fundamental paradigms. For instance, HotpotQA~\cite{Yang2018Hotpot}, a multi-hop question answering task, can be solved by combining \texttt{Matching} and \texttt{MRC}, where \texttt{Matching} is responsible for finding relevant documents and \texttt{MRC} is responsible for selecting the answer span~\cite{Wu2021Graph}.

\section{Paradigm Shift in NLP Tasks}
\label{sec:paradigm_shift}
In this section, we review the paradigm shifts that occur in different NLP tasks: Text Classification, Natural Language Inference, Named Entity Recognition, Aspect-Based Sentiment Analysis, Relation Exaction, Text Summarization, and Parsing.

\begin{sidewaystable*}
\centering
\resizebox{\linewidth}{!}{
    \begin{tabular}{ll|llllll}
        \toprule
        \multicolumn{2}{l|}{\textbf{Task}} & \textbf{Original Paradigm} & \multicolumn{5}{c}{\textbf{Shifted Paradigm}}                                                                                                                                  \\ \midrule
        \multirow{4}{*}{\textbf{TC}}      & Paradigm & \texttt{Class} & \multicolumn{1}{c}{\multirow{4}{*}{\Huge $\Rightarrow$}} & \multicolumn{1}{l|}{\texttt{Matching}} & \multicolumn{1}{l|}{\texttt{Seq2Seq}} & \multicolumn{1}{l|}{\texttt{(M)LM}}   &       \\
        & Input & $\cX$ & \multicolumn{1}{c}{} & \multicolumn{1}{l|}{$\cX,\cL$} & \multicolumn{1}{l|}{$\cX$} &\multicolumn{1}{l|}{$f_{prompt}(\cX)$} & \\
        & Output & $\cY$ & \multicolumn{1}{c}{} & \multicolumn{1}{l|}{$\cY\in\{0,1\}$} & \multicolumn{1}{l|}{$y_1,\cdots,y_m$} & \multicolumn{1}{l|}{$g(\cY)$} & \\
        & Example & \cite{devlin-etal-2019-bert} & \multicolumn{1}{c}{} & \multicolumn{1}{l|}{\cite{Chai20Description}} & \multicolumn{1}{l|}{\cite{Yang2018SGM}}        & \multicolumn{1}{l|}{\cite{Schick21PET}} & \\ \midrule
        \multirow{4}{*}{\textbf{NLI}}     & Paradigm & \texttt{Matching} & \multirow{4}{*}{\Huge $\Rightarrow$} & \multicolumn{1}{l|}{\texttt{Class}} & \multicolumn{1}{l|}{\texttt{Seq2Seq}} & \multicolumn{1}{l|}{\texttt{(M)LM}} & \\
        & Input & $\cX_a, \cX_b$ & & \multicolumn{1}{l|}{$\cX_a\oplus\cX_b$} & \multicolumn{1}{l|}{$f_{prompt}(\cX_a, \cX_b)$} & \multicolumn{1}{l|}{$f_{prompt}(\cX_a, \cX_b)$} & \\
        & Output & $\cY$ & & \multicolumn{1}{l|}{$\cY$} & \multicolumn{1}{l|}{$\cY$} & \multicolumn{1}{l|}{$g(\cY)$} &       \\
        & Example & \cite{Chen2017ESIM} & & \multicolumn{1}{l|}{\cite{devlin-etal-2019-bert}}         & \multicolumn{1}{l|}{\cite{McCann18decaNLP}} & \multicolumn{1}{l|}{\cite{Schick21PET}} &       \\ \midrule
        \multirow{4}{*}{\textbf{NER}}     & Paradigm & \texttt{SeqLab} & \multirow{4}{*}{\Huge $\Rightarrow$} & \multicolumn{1}{l|}{\texttt{Class}} & \multicolumn{1}{l|}{\texttt{MRC}} & \multicolumn{1}{l|}{\texttt{Seq2Seq}} & \texttt{(M)LM} \\
        & Input & $x_1,\cdots,x_n$ & & \multicolumn{1}{l|}{$\cX_{span}$}         & \multicolumn{1}{l|}{$\cX,\cQ_y$}        & \multicolumn{1}{l|}{$\cX$}        & $\cX$      \\
        & Output & $y_1,\cdots,y_n$ &                                                      & \multicolumn{1}{l|}{$\cY$}         & \multicolumn{1}{l|}{$\cX_{span}$}        & \multicolumn{1}{l|}{$(\cX_{ent_i},\cY_{ent_i})_{i=1}^m$}        & $g(\cY)$      \\
        & Example & \cite{Ma16End2end} & & \multicolumn{1}{l|}{\cite{fu2021spanner}}         & \multicolumn{1}{l|}{\cite{Li20MRC}}        & \multicolumn{1}{l|}{\cite{Yan2021UniNER}}        & \cite{Cui2021Template}      \\ \midrule
        \multirow{4}{*}{\textbf{ABSA}}    & Paradigm & \texttt{Class} & \multirow{4}{*}{\Huge $\Rightarrow$}                     & \multicolumn{1}{l|}{\texttt{Matching}} & \multicolumn{1}{l|}{\texttt{MRC}}     & \multicolumn{1}{l|}{\texttt{Seq2Seq}} & \texttt{(M)LM} \\
        & Input & $\cX_{asp}$ & & \multicolumn{1}{l|}{$\cX,\cS_{aux}$} & \multicolumn{1}{l|}{$\cX,\cQ_{asp},\cQ_{opin},\cQ_{sent}$} & \multicolumn{1}{l|}{$\cX$}        & $f_{prompt}(\cX)$      \\
        & Output & $\cY$ & & \multicolumn{1}{l|}{$\cY$} & \multicolumn{1}{l|}{$\cX_{asp}, \cX_{opin}, \cY_{sent}$} & \multicolumn{1}{l|}{$(\cX_{asp_i},\cX_{opin_i},\cY_{sent_i})_{i=1}^m$}        & $g(\cY)$      \\
        & Example  & \cite{Wang2016Attention} & & \multicolumn{1}{l|}{\cite{sun-etal-2019-utilizing}} & \multicolumn{1}{l|}{\cite{Mao2021Joint}} & \multicolumn{1}{l|}{\cite{Yan2021UniABSA}}        &  \cite{Li2021Sentiprompt}     \\ \midrule
        \multirow{4}{*}{\textbf{RE}}      & Paradigm & \texttt{Class} & \multirow{4}{*}{\Huge $\Rightarrow$}                     & \multicolumn{1}{l|}{\texttt{MRC}}      & \multicolumn{1}{l|}{\texttt{Seq2Seq}} & \multicolumn{1}{l|}{\texttt{(M)LM}}   &       \\
        & Input & $\cX$ & & \multicolumn{1}{l|}{$\cX,\cQ_y$}         & \multicolumn{1}{l|}{$\cX$}        & \multicolumn{1}{l|}{$f_{prompt}(\cX)$}        &       \\
        & Output & $\cY$ & & \multicolumn{1}{l|}{$\cX_{ent}$} & \multicolumn{1}{l|}{$(\cY_i,\cX_{sub_i},\cX_{obj_j})_{i=1}^m$}        & \multicolumn{1}{l|}{$g(\cY)$}        &       \\
        & Example & \cite{Zeng2014Relation} &                                                      & \multicolumn{1}{l|}{\cite{Levy2017Zero}}         & \multicolumn{1}{l|}{\cite{Zeng2018Extracting}}        & \multicolumn{1}{l|}{\cite{Han2021PTR}}        &       \\ \midrule
        \multirow{5}{*}{\textbf{Summ}}    & Paradigm & \texttt{SeqLab} / \texttt{Seq2Seq} & \multirow{4}{*}{\Huge $\Rightarrow$}                     & \multicolumn{1}{l|}{\texttt{Matching}} & \multicolumn{1}{l|}{\texttt{(M)LM}}   & \multicolumn{1}{l|}{}        &       \\
        & Input & $\cX_1,\cdots,\cX_n$ / $\cX, \cQ_{summ}$ & & \multicolumn{1}{l|}{$(\cX, \cS_{cand_i})_{i=1}^n$}         & \multicolumn{1}{l|}{$\cX, \mathrm{Keywords/Prompt}$} & \multicolumn{1}{l|}{}        &       \\
        & Output & $\cY_1,\cdots, \cY_n \in \{0,1\}^n$ / $\cY$ & & \multicolumn{1}{l|}{$\hat{\cS}_{cand}$} & \multicolumn{1}{l|}{$\cY$}        & \multicolumn{1}{l|}{}        &       \\
        & Example & \cite{Cheng2016Summ} & & \multicolumn{1}{l|}{\cite{zhong-etal-2020-extractive}} & \multicolumn{1}{l|}{\cite{Aghajanyan2021HTML}}        & \multicolumn{1}{l|}{}        &       \\
        & & \cite{McCann18decaNLP} &                                                      & \multicolumn{1}{l|}{}         & \multicolumn{1}{l|}{}        & \multicolumn{1}{l|}{}        &       \\
        \midrule
        \multirow{4}{*}{\textbf{Parsing}} & Paradigm & \texttt{Seq2ASeq} & \multirow{4}{*}{\Huge $\Rightarrow$} & \multicolumn{1}{l|}{\texttt{(M)LM}} & \multicolumn{1}{l|}{\texttt{SeqLab}}  & \multicolumn{1}{l|}{\texttt{MRC}}  & \texttt{Seq2Seq} \\
        & Input & $(\cX,\cC_t)_{t=0}^{m-1}$ & & \multicolumn{1}{l|}{$(\cX, \cY_i)_{i=1}^k$} & \multicolumn{1}{l|}{$x_1,\cdots,x_n$}        & \multicolumn{1}{l|}{$\cX,\cQ_{child}$}        &  $\cX$     \\
        & Output & $\cA = a_1,\cdots,a_m$ & & \multicolumn{1}{l|}{$\hat{\cY}$}         & \multicolumn{1}{l|}{$g(y_1,\cdots,y_n)$}        & \multicolumn{1}{l|}{$\cX_{parent}$}        &$g(y_1,\cdots,y_m)$       \\
        & Example & \cite{Chen2014Fast} & & \multicolumn{1}{l|}{\cite{Choe2016Parsing}}         & \multicolumn{1}{l|}{\cite{Strzyz2019Viable}}        & \multicolumn{1}{l|}{\cite{Gan21Dependency}}        & \cite{Vinyals2015Grammar}      \\ \bottomrule
        \end{tabular}
}
\caption{
    Paradigms shift in NLP tasks. \textbf{TC}: text classification. \textbf{NLI}: natural language inference. \textbf{NER}: named entity recognition. \textbf{ABSA}: aspect-based sentiment analysis. \textbf{RE}: relation extraction. \textbf{Summ}: text summarization. \textbf{Parsing}: syntactic/semantic parsing. $f$ and $g$ indicate pre-processing and post-processing, respectively. 
    $\cL$ means label description. $\oplus$ means concatenation. $\cX_{asp}, \cX_{opin}, \cY_{sent}$ mean aspect, opinion, and sentiment, respectively. $\cS_{aux}$ means auxiliary sentence. $\cX_{sub}, \cX_{obj}$ stand for subject entity and object entity, respectively. $\cS_{cand}$ means candidate summary. $\cC_t$ is the configuration at time step $t$ and $\cA$ is a sequence of actions.}
% Paradigms shift in natural language processing tasks. \textbf{TC}: text classification. \textbf{NLI}: natural language inference. \textbf{NER}: named entity recognition. \textbf{ABSA}: aspect-based sentiment analysis. \textbf{RE}: relation extraction. \textbf{Summ}: text summarization. \textbf{Parsing}: syntactic/semantic parsing. $f$ and $g$ indicate pre-processing and post-processing, respectively. In \texttt{(M)LM}, $f(\cdot)$ is usually implemented as a template and $g(\cdot)$ is a verbalizer. In parsing tasks, $g(\cdot)$ is a function that reconstructs the structured representation (\textit{e.g.} dependency tree) from the output sequence.
% $\cL$ means label description. $\oplus$ means concatenation. $\cX_{asp}, \cX_{opin}, \cY_{sent}$ mean aspect, opinion, and sentiment, respectively. $\cS_{aux}$ means auxiliary sentence. $\cX_{sub}, \cX_{obj}$ stand for subject entity and object entity, respectively. $\cS_{cand}$ means candidate summary. $\cC_t$ is configuration $t$ and $\cA$ is a sequence of actions. More details can be found in Section~\ref{sec:paradigm_shift}.}
\label{tb:paradigmshift}
\end{sidewaystable*}

\subsection{Text Classification}
Text classification is an essential task in various NLP applications. Conventional text classification tasks can be well solved by the \texttt{Class} paradigm. Nevertheless, its variants such as multi-label classification can be challenging, in which case \texttt{Class} may be sub-optimal. To that end, \citet{Yang2018SGM} propose to adopt the \texttt{Seq2Seq} paradigm to better capture interactions between the labels for multi-label classification tasks.

In addition, the semantics hidden in the labels can not be fully exploited in the \texttt{Class} paradigm. \citet{Chai20Description} and \citet{Wang21Entailment} adopt the \texttt{Matching} paradigm to predict whether the pair-wise input $(\cX, \cL_y)$ is matched, where $\cX$ is the original text and $\cL_y$ is the label description for class $y$. Though the semantic meaning of a label can be exactly defined by the samples belonging to it, incorporating prior knowledge of the label is also helpful when training data is limited.

As the rise of pre-trained language models (LMs), text classification tasks can also be solved in the \texttt{(M)LM} paradigm~\cite{Brown2020GPT3,Schick21PET,Schick21Size,Gao20Making}. By reformulating a text classification task into a (masked) language modeling task, the gap between LM pre-training and fine-tuning is narrowed, resulting in improved performance when training data is limited.

\subsection{Natural Language Inference}
Natural Language Inference (NLI) is typically modeled in the \texttt{Matching} paradigm, where the two input texts $(\cX_a, \cX_b)$ are encoded and interact with each other, followed by a classifier to predict the relationship between them~\cite{Chen2017ESIM}. With the emergence of powerful encoder such as BERT~\cite{devlin-etal-2019-bert}, NLI tasks can be simply solved in the \texttt{Class} paradigm by concatenating the two texts as one. In the case of few-shot learning, NLI tasks can also be formulated in the \texttt{(M)LM} paradigm by modifying the input, \textit{e.g.} "$\cX_a$ ? \texttt{[MASK]} , $\cX_b$". The unfilled token \texttt{[MASK]} can be predicted by the MLM head as Yes/No/Maybe, corresponding to Entailment/Contradiction/Neutral~\cite{Schick21PET,Schick21Size,Gao20Making}.

\subsection{Named Entity Recognition}
Named Entity Recognition (NER) is also a fundamental task in NLP. NER can be categorized into three subtasks: flat NER, nested NER, and discontinuous NER. Traditional methods usually solve the three NER tasks based on three paradigms respectively, \textit{i.e.} \texttt{SeqLab}~\cite{Ma16End2end,Lample2016NeuralNER}, \texttt{Class}~\cite{Xia2019MultigrainedNER,Fisher2019MergeLabel}, and \texttt{Seq2ASeq}~\cite{Lample2016NeuralNER,Dai2020TransitionNER}.
% In recent years, there has been work to shift from \texttt{SeqLab} to other paradigms (\textit{e.g.} \texttt{Class}, \texttt{MRC} and \texttt{Seq2Seq}).

\citet{yu-etal-2020-named} and \citet{fu2021spanner} solve flat NER and nested NER with the \texttt{Class} paradigm. The main idea is to predict the label for each span in the input text. This paradigm shift introduces the span overlapping problem: The predicted entities may be overlapped, which is not allowed in flat NER. To handle this, \citet{fu2021spanner} adopt a heuristic decoding method: For these overlapped spans, only keep the span with the highest prediction probability.

% For the Flat Ner task, they used their own methods to avoid the prediction of overlapped spans. For example, the Heuristic Decoding method used in the framework of \citet{fu2021spanner}, only keeps the span with the highest prediction probability from a span $S_i$ and all of his sub-spans $[S_{i,1}, ..., S_{i,j}]$

\citet{Li20MRC} propose to formulate flat NER and nested NER as a MRC task. They reconstruct each sample into a triplet $(\cX, \cQ_y, \cX_{span})$, where $\cX$ is the original text, $\cQ_y$ is the question for entity $y$, $\cX_{span}$ is the answer. Given context, question, and answer, the \texttt{MRC} paradigm can be adopted to solve this. Since there can be multiple answers (entities) in a sentence, an index matching module is developed to align the start and end indexes.

% $(q_y, x_{start,end}, X)$ based on the original sentences and the corresponding golden labels, and the elements in the triplet correspond to Question, Answer and Context required by the MRC task respectively. In this paper, authors constructed a related question for each tag category based on the dataset annotation guidelines. According to the paradigm of the MRC task, authors concatenated the question and the original text as input to the model, and then predicting where the entities start and end based on the model output. Because it is possible to get multiple start and end indexes of one entity, an index matching module is introduced to align the start and end indexes.

\citet{Yan2021UniNER} use a unified model based on the \texttt{Seq2Seq} paradigm to solve all the three kinds of NER subtasks. The input of the \texttt{Seq2Seq} paradigm is the original text, while the output is a sequence of span-entity pairs, for instance, "\textit{Barack Obama} \texttt{<Person>} \textit{US} \texttt{<Location>}". Due to the versatility of the \texttt{Seq2Seq} paradigm and the great power of BART~\cite{bart}, this unified model achieved state-of-the-art performance on various datasets spanning all the three NER subtasks.

% They represented an entity as a sequence $[s_{i1}, e_{i1}, ..., s_{ij}, e_{ij}, t_i]$, where $s_{i*}$ and $e_{i*}$ are the start and end index of a span, respectively. Since an entity may contain multiple spans (\textit{e.g.} discontinuous entity), the entity sequence representation here contains $j$ spans. $t_i$ is the entity tag index. The sequences of all entities in the original sample are spliced together to produce the final target sequence. The pre-training model BART \cite{bart} was used to solve the NER tasks under the Seq2Seq paradigm. In order not to confuse token index  in the sample and entity tag index, authors specifically introduced token index predicting and Tag index predicting modules to obtain token index distribution and tag index distribution. Finally, the two distributions are spliced together to perform the prediction of the current time step.

\subsection{Aspect-Based Sentiment Analysis}
%classification, seq2seq, matching
Aspect-Based Sentiment Analysis (ABSA) is a fine-grained sentiment analysis task with seven subtasks, \textit{i.e.}, Aspect Term Extraction (AE), Opinion Term Extraction (OE), Aspect-Level Sentiment Classification (ALSC), Aspect-oriented Opinion Extraction (AOE), Aspect Term Extraction and Sentiment Classification (AESC), Pair Extraction (Pair), and Triplet Extraction (Triplet). These subtasks can be solved by different paradigms. For example, ALSC can be solved by the \texttt{Class} paradigm, and AESC can be solved by the \texttt{SeqLab} paradigm.

ALSC is to predict the sentiment polarity for each target-aspect pair, \textit{e.g.} (\texttt{LOC1}, price), given a context, \textit{e.g.} "\texttt{LOC1} \textit{is often considered the coolest area of London}". \citet{sun-etal-2019-utilizing} formulate such a classification task into a sentence-pair matching task, and adopt the \texttt{Matching} paradigm to solve it. In particular, they generate auxiliary sentences (denoted as $\cS_{aux}$) for each target-aspect pair. For example, $\cS_{aux}$ for (\texttt{LOC1}, price) can be "\textit{What do you think of the price of} \texttt{LOC1}?". The auxiliary sentence is then concatenated with the context as $(\cS_{aux}, \cX)$, which is then fed into BERT~\cite{devlin-etal-2019-bert} to predict the sentiment.

% The author constructed $C$ sentences according to the given aspect in the sample and $C$ sentiment polarities in the dataset, and then judged the polarity of aspect in the sample through the matching score of the original sample and the $C$ sentences.

\citet{Mao2021Joint} adopt the \texttt{MRC} paradigm to handle all of the ABSA subtasks. In particular, they construct two queries to sequentially extract the aspect terms and their corresponding polarities and opinion terms. The first query is "\textit{Find the aspect terms in the text.}" Assume the answer (aspect term) predicted by the MRC model is \texttt{AT}, then the second query can be constructed as "\textit{Find the sentiment polarity and opinion terms for} \texttt{AT} \textit{in the text.}" Through such dataset conversion, all ABSA subtasks can be solved in the \texttt{MRC} paradigm.

% designed a method to solve all ABSA subtasks in the paradigm of MRC. For all ABSA subtasks, the most difficult is Triplet Extraction, that is, to extract aspects, their corresponding opinion terms and sentiment polarity at the same time. Because this subtask can be said to be a combination of other subtasks. The authors divided the Triplet Extraction task into AE, SC and AOE tasks, and used a parameter sharing model to solve these three tasks simultaneously. For AE tasks, the authors constructed a query: “\textit{Find the aspect terms in the text.}”; For SC and AOE tasks, the authors constructed a query: “\textit{Find the sentiment polarity and opinion terms for AT in the text.}”. This is why the Dual-MRC is called in the paper. To solve different ABSA subtasks, we only need different combinations of AE, SC and AOE tasks.

\citet{Yan2021UniABSA} solve all the ABSA subtasks with the \texttt{Seq2Seq} paradigm by converting the original label of a subtask into a sequence of tokens, which is used as the target to train a seq2seq model. Take the Triplet Extraction subtask as an example, for a input sentence, "\textit{The \textcolor{red}{drinks} are always \textcolor{blue}{well made} and \textcolor{red}{wine selection} is \textcolor{blue}{fairly priced}}", the output target is constructed as "\textit{\textcolor{red}{drinks} \textcolor{blue}{well made}} \texttt{Positive} \textit{\textcolor{red}{wine selection} \textcolor{blue}{fairly priced}} \texttt{Positive}". Equipped with BART~\cite{bart} as the backbone, they achieved competitive performance on most ABSA subtasks.

Very recently, \citet{Li2021Sentiprompt} propose to formulate the ABSA subtasks in the \texttt{(M)LM} paradigm. In particular, for the input text $\cX$, and the aspect $A$ and opinion $O$ of interest, they construct a consistency prompt and a polarity prompt as: \textit{The $A$ is $O$?} \texttt{[MASK]}. \textit{This is} \texttt{[MASK]}, where the first \texttt{[MASK]} can be filled with \textit{yes} or \textit{no} for consistent or inconsistent $A$ and $O$, and the second \texttt{[MASK]} can be filled with sentiment polarity words.

% also proposed a method that can solve all ABSA subtasks. The authors converted all tasks into the Seq2Seq paradigm. They designed a corresponding target sequence for each subtask. For example, for the Triplet Extraction task, the target sequence $Y$ can be expressed as $[a_1^s, a_1^e, o_1^s, o_1^e, s_1^p, ..., a_i^s, a_i^e, o_i^s, o_i^e, s_i^p, ...]$, where $a_*^s$ and $a_*^e$ respectively represent the start and end index of an aspect, $o_*^s$ and $o_*^e$ respectively represent the start and end index of the opinion term corresponding to aspect $(a_*^s - a_*^e)$, and  $s_*^p$ represents the index of sentiment polarity class.

\begin{figure*}[t!]
    \centering
    \includegraphics[width=\linewidth]{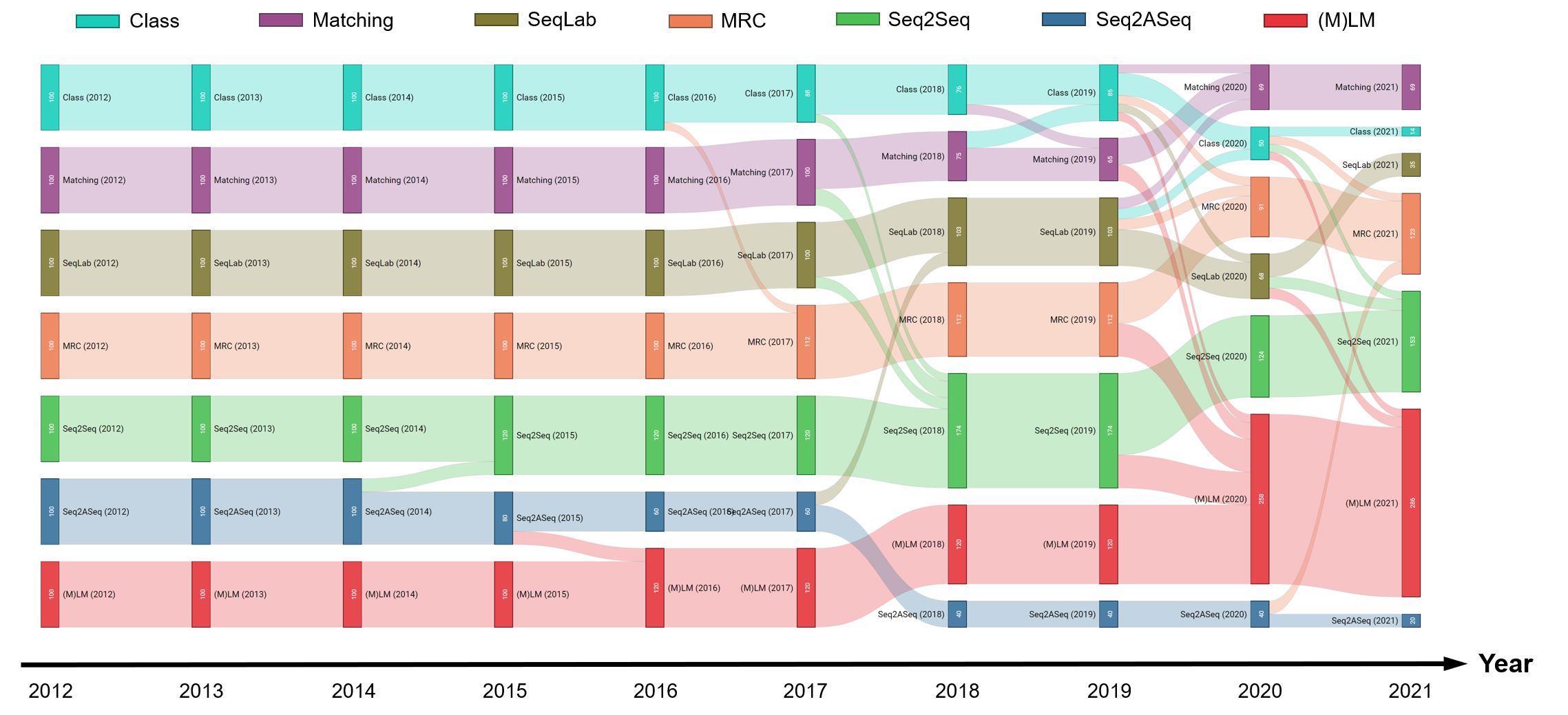}
    \caption{Sankey diagram to depict the trend of paradigm shifting and unifying in natural language processing tasks. In Section~\ref{sec:trends} we show how this diagram is drawn.}
    \label{fig:paradigm_shift}
\end{figure*}

\subsection{Relation Exaction}
%SeqLab
Relation Extraction (RE) has two main subtasks: Relation Prediction (predicting the relationship $r$ of two given entities $s$ and $o$ conditioned on their context) and Triplet Extraction (extracting triplet $(s,r,o)$ from the input text). The former subtask is mainly solved with the \texttt{Class} paradigm~\cite{Zeng2014Relation,Sun2020Colake}, while the latter subtask is often solved in the pipeline style that first uses the \texttt{SeqLab} paradigm to extract the entities and then uses the \texttt{Class} paradigm to predict the relationship between the entities. Recent years have seen paradigm shift in relation extraction, especially in triplet extraction.

\citet{Zeng2018Extracting} solve the triplet extraction task with the \texttt{Seq2Seq} paradigm. In their framework, the input of the \texttt{Seq2Seq} paradigm is the original text, while the output is a sequence of triplets $\{(r_1,s_1,o_1), \cdots (r_n,s_n,o_n)\}$. The copy mechanism~\cite{Gu2016Copy} is adopted to extract entities in the text.

\citet{Levy2017Zero} address the RE task via the \texttt{MRC} paradigm by generating relation-specific questions. For instance, for relation $educated\_at(s,o)$, a question such as "\textit{Where did} $s$ \textit{graduate from?}" can be crafted to query a MRC model. Moreover, they demonstrate that formulating the RE task with \texttt{MRC} has a potential of zero-shot generalization to unseen relation types. Further, \citet{Li19Entity} and \citet{Zhao2020Asking} formulate the triplet extraction task as multi-turn question answering and solve it with the \texttt{MRC} paradigm. They extract entities and relations from the text by progressively asking the MRC model with different questions.

Very recently, \citet{Han2021PTR} formulate the RE task as a MLM task by using logic rules to construct prompts with multiple sub-prompts. By encoding prior knowledge of entities and relations into prompts, their proposed model, PTR, achieved state-of-the-art performance on multiple RE datasets.

\subsection{Text Summarization}
Text Summarization aims to generate a concise and informative summary of large texts. There are two different approaches to solve the text summarization task: Extractive Summarization and Abstractive Summarization. Extractive summarization approaches extract the clauses of the original text to form the final summary, which usually lies in the \texttt{SeqLab} paradigm. In contrast, abstractive summarization approaches usually adopt the \texttt{Seq2Seq} paradigm to directly generate a summary conditioned on the original text.

\citet{McCann18decaNLP} reformulate the summarization task as a question answering task, where the question is "\textit{What is the summary?}". Since the answer (\textit{i.e.} the summary) is not necessarily comprised of the tokens in the original text, traditional MRC model cannot handle this. Therefore, the authors developed a seq2seq model to solve the summarization task in such format.

% The authors proposed a multi-task framework, in which ten tasks including Summarization tasks are unified in the MRC paradigm. They generated questions for a sample of each task, which they then reconstructed into a new sample containing context, questions, and answers. When generating answers, the framework not only ensures that the generated words can be directly derived from the contexts or questions of the input, but also generates words from the output vocabulary.

\citet{zhong-etal-2020-extractive} propose to solve the extractive summarization task in the \texttt{Matching} paradigm instead of the \texttt{SeqLab} paradigm. The main idea is to match the semantics of the original text and each candidate summary, finding the summary with the highest matching score. Compared with traditional methods of extracting sentences individually, the matching framework enables the summary extractor to work at summary level rather than sentence level.

% \citet{He2020CTRLSum} pre-train a controllable summarization model, CTRLsum, which allows users to control the generated summary by keywords and prompts.

\citet{Aghajanyan2021HTML} formulate the text summarization task in the \texttt{(M)LM} paradigm. They pre-train a BART-style model directly on large-scale structured HTML web pages. Due to the rich semantics encoded in the HTML keywords, their pre-trained model is able to perform zero-shot text summarization by predicting the \texttt{<title>} element given the \texttt{<body>} of the document.

% The main idea is that some candidate summaries have higher ROUGE \cite{lin-hovy-2003-automatic} score with golden summary at the summary level (considers sentences in candidate summaries as a whole and then calculates the ROUGE score with the golden summary), but lower ROUGE score with golden summary at the sentence level (indicates the average overlaps between each sentence in candidate summaries and the golden summary). Therefore, it is difficult for sentence-level extractors to extract such summaries well. Based on the BERT model \cite{devlin-etal-2019-bert}, the authors constructed Siamese-Bert inspired by the siamese neural network structure to match the original document and candidate summary from the summary level.

\begin{table*}[h]
    \centering
    \resizebox{\linewidth}{!}{
    \begin{tabular}{l|c|c|c|c}
    \toprule
        \textbf{Year} & \textbf{Task} & \textbf{Original Paradigm} & \textbf{Shifted Paradigm} & \textbf{Paper} \\
    \midrule
2015          & Parsing                            & \texttt{Seq2ASeq}  & \texttt{Seq2Seq}   &              \cite{Vinyals2015Grammar}  \\
2016          & Parsing                            & \texttt{Seq2ASeq}  & \texttt{(M)LM}                    &    \cite{Choe2016Parsing}            \\
2017          & Relation Extraction                & \texttt{Class}     & \texttt{MRC}       &            \cite{Levy2017Zero}    \\
2018          & Text Summarization                 & \texttt{SeqLab}    & \texttt{Seq2Seq}       &            \cite{McCann18decaNLP}    \\
2018          & Parsing                            & \texttt{Seq2ASeq}  & \texttt{SeqLab}    &            \cite{Gomez2018Constituent}    \\
2018          & Natural Language Inference         & \texttt{Matching}  & \texttt{Seq2Seq}       &             \cite{McCann18decaNLP}   \\
% 2018          & Text Classification & \texttt{Class}     & \texttt{MRC}       &     \cite{McCann18decaNLP}           \\
% 2018          & Relation Extraction                & \texttt{Class}     & \texttt{MRC}       &   \cite{McCann18decaNLP}             \\
2018          & Text Classification & \texttt{Class}     & \texttt{Seq2Seq}   &     \cite{McCann18decaNLP}          \\
2018          & Relation Extraction                & \texttt{Class}     & \texttt{Seq2Seq}   &  \cite{Zeng2018Extracting}\\
2019          & Sentiment Analysis                 & \texttt{Class}     & \texttt{Matching}  & \cite{sun-etal-2019-utilizing}\\
2019          & Natural Language Inference         & \texttt{Matching}  & \texttt{Class}     & \cite{devlin-etal-2019-bert}\\
2020          & Named Entity Recognition           & \texttt{SeqLab}    & \texttt{Class}     & \cite{yu-etal-2020-named}  \\
2020          & Named Entity Recognition           & \texttt{SeqLab}    & \texttt{MRC}       & \cite{Li20MRC}               \\
2020          & Text Summarization                 & \texttt{SeqLab}    & \texttt{Matching}  & \cite{zhong-etal-2020-extractive}               \\
2020          & Event Extraction                   & \texttt{Class}     & \texttt{MRC}       & \cite{Liu2020Event}               \\
2020          & Event Extraction                   & \texttt{Class}     & \texttt{SeqLab}    & \cite{Ramponi2020Event}               \\
2020          & Text Classification & \texttt{Class}     & \texttt{Matching}  & \cite{Yin20Universal}               \\
2020          & Text Classification & \texttt{Class}     & \texttt{(M)LM}                    & \cite{Brown2020GPT3}               \\
2020          & Question Answering                 & \texttt{MRC}       & \texttt{(M)LM}                    &  \cite{Brown2020GPT3}              \\
2020          & Machine Translation                & \texttt{Seq2Seq}   & \texttt{(M)LM}                    &  \cite{Brown2020GPT3}              \\
2020          & Natural Language Inference         & \texttt{Matching}  & \texttt{(M)LM}                    &  \cite{Brown2020GPT3}              \\
2021          & Named Entity Recognition           & \texttt{SeqLab}    & \texttt{Seq2Seq}   &  \cite{Yan2021UniNER}              \\
2021          & Named Entity Recognition           & \texttt{SeqLab}  & \texttt{(M)LM}       &  \cite{Cui2021Template}         \\    
2021          & Sentiment Analysis                 & \texttt{Class}     & \texttt{MRC}       &   \cite{Mao2021Joint}             \\
2021          & Sentiment Analysis                 & \texttt{Class}     & \texttt{Seq2Seq}   &  \cite{Yan2021UniABSA}              \\
2021          & Sentiment Analysis                 & \texttt{Class}  & \texttt{(M)LM}       &  \cite{Schick21PET}         \\    
2021          & Parsing                            & \texttt{Seq2ASeq}  & \texttt{MRC}       &  \cite{Gan21Dependency}         \\    
    \bottomrule
    \end{tabular}
    }
    \caption{Source data of Figure~\ref{fig:paradigm_shift}. We only list the first work for each paradigm shift.}
    \label{tab:source_data}
\end{table*}

\subsection{Parsing}
Parsing (constituency parsing, dependency parsing, semantic parsing, \textit{etc}.) plays a crucial role in many NLP applications such as machine translation and question answering. This family of tasks is to derive a structured syntactic or semantic representation from a natural language utterance. Two commonly used approaches for parsing are transition-based methods and graph-based methods. Typically, transition-based methods lie in the \texttt{Seq2ASeq} paradigm, and graph-based methods lie in the \texttt{Class} paradigm.

By linearizing the target tree-structure to a sequence, parsing can be solved in the \texttt{Seq2Seq} paradigm~\cite{Andreas2013Semantic,Vinyals2015Grammar,Li2018Seq2seq,Rongali2020Dontparse}, the \texttt{SeqLab} paradigm~\cite{Gomez2018Constituent,Strzyz2019Viable,Vilares2020Discontinuouts,Vacareanu2020Parsing}, and the \texttt{(M)LM} paradigm~\cite{Choe2016Parsing}. In addition, \citet{Gan21Dependency} employ the \texttt{MRC} paradigm to extract the parent span given the original sentence as the context and the child span as the question, achieving state-of-the-art performance on dependency parsing tasks across various languages.

\subsection{Trends of Paradigm Shift}
\label{sec:trends}
% To intuitively depict the trend of paradigm shift, we draw a sankey diagram\footnote{Sankey diagram is a visualization used to depict data flows. Our sankey diagram is generated by \href{http://sankey-diagram-generator.acquireprocure.com}{http://sankey-diagram-generator.acquireprocure.com}.} in Figure~\ref{fig:paradigm_shift}. We track the development of the NLP tasks considered in this section, along with several additional common tasks such as event extraction. When a task is solved by a paradigm that is different with its original paradigm, some of the values is transferred from the original paradigm to the new paradigm. We initialize the value of each paradigm as 100, and the transferred value is defined as $100/N$, where $N$ is the total number of paradigms that have been used to solve the task. Therefore, each branch in Figure~\ref{fig:paradigm_shift} indicates a task that shifts its paradigm.

To intuitively depict the trend of paradigm shifts, we draw a Sankey diagram\footnote{Sankey diagram is a visualization used to depict data flows. Our sankey diagram is generated by \href{http://sankey-diagram-generator.acquireprocure.com}{http://sankey-diagram-generator.acquireprocure.com}.} in Figure~\ref{fig:paradigm_shift}. We track the development of the NLP tasks considered in this section, along with several additional common tasks such as event extraction. When a task is solved using a paradigm that is different from its original paradigm, some of the values of the original paradigm are transferred to the new paradigm. In particular, for each NLP task of interest, we collect published papers that solve this task from 2012 to 2021 and denote the paradigm used in 2012 as the original paradigm of this task. Then we track the paradigm shifts in all the tasks with the same original paradigm and count the number of tasks that observed paradigm shifts until 2021. For each paradigm, we denote $N$ as the total number of tasks that branched out to new paradigms. Assume that the initial value of each paradigm is 100, and the transferred value for each out-branch is defined as $100/(N+1)$. Therefore, each branch in Figure~\ref{fig:paradigm_shift} indicates a task that shifted its paradigm. Table~\ref{tab:source_data} lists the source data of Figure~\ref{fig:paradigm_shift}.

As shown in Figure~\ref{fig:paradigm_shift}, we find that: (1) The frequency of paradigm shift is increasing in recent years, especially after the emergence of pre-trained language models (PTMs).
To fully utilize the power of these PTMs, a better way is to reformulate various NLP tasks into the paradigms that PTMs are good at.
(2) More and more NLP tasks have shifted from traditional paradigms such as \texttt{Class}, \texttt{SeqLab}, \texttt{Seq2ASeq}, to paradigms that are more general and flexible, \textit{i.e.}, \texttt{(M)LM}, \texttt{Matching}, \texttt{MRC}, and \texttt{Seq2Seq}, which will be discussed in the following section.

%transition-based, seqlab, seq2seq

% \subsection{\texttt{Matching}~$\rightarrow$~\texttt{Class}}

% \subsection{\texttt{Class}~$\rightarrow$~\texttt{Matching}}

% \subsection{\texttt{Class}~$\rightarrow$~\texttt{Seq2Seq}}

% \subsection{\texttt{SeqLab}~$\rightarrow$~\texttt{Matching}}

% \subsection{\texttt{SeqLab}~$\rightarrow$~\texttt{MRC}}

% \subsection{\texttt{SeqLab}~$\rightarrow$~\texttt{Seq2Seq}}

\section{Potential Unified Paradigms in NLP}
\label{sec:unified_paradigm}
Some of the paradigms have demonstrated potential ability to formulate various NLP tasks into a unified framework. Instead of solving each task separately, such paradigms provide the possibility that a single deployed model can serve as a unified solver for diverse NLP tasks. The advantages of a single unified model over multiple task-specific models can be summarized as follows:
\begin{itemize}
    \item \textbf{Data efficiency.} Training task-specific models usually requires large-scale task-specific labeled data. In contrast, unified model has shown its ability to achieve considerable performance with much less labeled data.
    \item \textbf{Generalization.} Task-specific models are hard to transfer to new tasks while unified model can generalize to unseen tasks by formulating them into proper formats.
    \item \textbf{Convenience.} The unified models are easier and cheaper to deploy and serve, making them favorable as commercial black-box APIs.
\end{itemize}

In this section, we discuss the following general paradigms that have the potential to unify diverse NLP tasks: \texttt{(M)LM}, \texttt{Matching}, \texttt{MRC}, and \texttt{Seq2Seq}.

\subsection{\texttt{(M)LM}}
Reformulating downstream tasks into a (M)LM task is a natural way to utilizing the pre-trained LMs. The original input is modified with a pre-defined or learned \textit{prompt} with some unfilled slots, which can be filled by the pre-trained LMs. Then the task labels can be derived from the filled tokens. For instance, a movie review "\textit{I love this movie}" can be modified by appending a prompt as "\textit{I love this movie. It was} \texttt{[MASK]}", in which \texttt{[MASK]} may be predicted as "\textit{fantastic}" by the LM. Then the word "\textit{fantastic}" can be mapped to the label "\textit{positive}" by a \textit{verbalizer}. Solving downstream tasks in the \texttt{(M)LM} paradigm is also referred to \textit{prompt-based learning}. By fully utilizing the pre-trained parameters of the MLM head instead of training a classification head from scratch, prompt-based learning has demonstrated great power in few-shot and even zero-shot settings~\cite{Scao21How}.

% \subsubsection{Design Considerations}
% There are two main design considerations in the framework: prompt and verbalizer.

\paragraph{Prompt.} The choice of prompt is critical to the performance of a particular task. A good prompt can be \textbf{(1) Manually designed}. \citet{Brown2020GPT3,Schick21PET,Schick21Size} manually craft task-specific prompts for different tasks. Though it is heuristic and sometimes non-intuitive, hand-crafted prompts already achieved competitive performance on various few-shot tasks. \textbf{(2) Mined from corpora}. \citet{Jiang20How} construct prompts for relation extraction by mining sentences with the same subject and object in the corpus. \textbf{(3) Generated by paraphrasing}. \citet{Jiang20How} use back translation to paraphrase the original prompt into multiple new prompts. \textbf{(4) Generated by another pre-trained language model}. \citet{Gao20Making} generate prompts using T5~\cite{Raffel20T5} since it is pre-trained to fill in missing spans in the input. \textbf{(5) Learned by gradient descent}. \citet{Shin20Autoprompt} automatically construct prompts based on gradient-guided search. If prompt is not necessarily discrete, it can be optimized efficiently in continuous space. Recent works~\cite{Li20Prefix,Qin21Learning,Hambardzumyan20WARP,Liu21PTuning,Zhong21OptiPrompt} have shown that continuous prompts can also achieve competitive or even better performance.
% \begin{enumerate}
%     \item \textbf{Manually designed}, \cite{Schick21PET,Schick21Size}
%     \item \textbf{Mined from corpora} \cite{Jiang20How},
%     \item \textbf{Generated by paraphrasing} \cite{Jiang20How},
%     \item \textbf{Generated by another PLM} \cite{Gao20Making},
%     \item \textbf{Learned by gradient descent}. concrete~\cite{Shin20Autoprompt} / continuous~\cite{Li20Prefix,Qin21Learning,Hambardzumyan20WARP,Liu21PTuning,Zhong21OptiPrompt}.
% \end{enumerate}
%Ensemble.

\paragraph{Verbalizer.} The design of verbalizer also has a strong influence on the performance of prompt-based learning~\cite{Gao20Making}. A verbalizer can be \textbf{(1) Manually designed}. \citet{Schick21PET} heuristically designed verbalizers for different tasks and achieved competitive results. However, it is not always intuitive for many tasks (\textit{e.g.}, when class labels not directly correspond to words in the vocabulary) to manually design proper verbalizers. \textbf{(2) Automatically searched} on a set of labelled data~\cite{Schick20Auto,Gao20Making,Shin20Autoprompt,Liu21PTuning}. \textbf{(3) Constructed and refined with knowledge base}~\cite{Hu21Knowledgeable}.
% \begin{enumerate}
%     \item \textbf{Manually designed} \cite{Schick21PET},
%     \item \textbf{Automatically searched} \cite{Schick20Auto,Gao20Making,Shin20Autoprompt,Liu21PTuning},
%     \item \textbf{Constructed and refined with knowledge base} \cite{Hu21Knowledgeable}.
% \end{enumerate}
% Ensemble.

\paragraph{Parameter-Efficient Tuning}
Compared with fine-tuning where all model parameters need to be tuned for each task, prompt-based tuning is also favorable in its parameter efficiency. Recent study~\cite{Lester21Power} has demonstrated that tuning only prompt parameters while keeping the backbone model parameters fixed can achieve comparable performance with standard fine-tuning when models exceed billions of parameters. Due to the parameter efficiency, prompt-based tuning is a promising technique for the deployment of large-scale pre-trained LMs. \textbf{In traditional fine-tuning}, the server has to maintain a task-specific copy of the entire pre-trained LM for each downstream task, and inference has to be performed in separate batches. \textbf{In prompt-based tuning}, only a single pre-trained LM is required, and different tasks can be performed by modifying the inputs with task-specific prompts. Besides, inputs of different tasks can be mixed in the same batch, which makes the service highly efficient.\footnote{The reader is referred to \citet{Liu21prompt} for a more comprehensive survey of prompt-based learning.}

\subsection{\texttt{Matching}}
Another potential unified paradigm is \texttt{Matching}, or more specifically textual entailment (a.k.a. natural language inference). Textual entailment is the task of predicting two given sentences, premise and hypothesis: whether the premise entails the hypothesis, contradicts the hypothesis, or neither. Almost all text classification tasks can be reformulated as a textual entailment one~\cite{Dagan05PASCAL,Poliak18Collecting,Yin20Universal,Wang21Entailment}. For example, a labeled movie review \{$x$: \textit{I love this movie}, $y$: \textit{positive}\} can be modified as \{$x$: \textit{I love this movie} \texttt{[SEP]} \textit{This is a great movie}, $y$: \textit{entailment}\}. Similar to pre-trained LMs, entailment models are also widely accessible. Such universal entailment models can be pre-trained LMs that are fine-tuned on some large-scale annotated entailment datasets such as MNLI~\cite{Williams18MNLI}. In addition to obtaining the entailment model in a supervised fashion, \citet{Sun2021NSPBERT} show that the next sentence prediction head of BERT, without training on any supervised entailment data, can also achieve competitive performance on various zero-shot tasks.

\paragraph{Domain Adaptation}
The entailment model may be biased to the source domain, resulting in poor generalization to target domains. To mitigate the domain difference between the source task and the target task, \citet{Yin20Universal} propose the cross-task nearest neighbor module that matches instance representations and class representations in the source domain and the target domain, such that the entailment model can generalize well to new NLP tasks with limited annotations.

\paragraph{Label Descriptions}
For single sentence classification tasks, label descriptions for each class are required to be concatenated with the input text to be predicted by the entailment model. Label descriptions can be regarded as a kind of prompt to trigger the entailment model. \citet{Wang21Entailment} show that hand-crafted label descriptions with minimum domain knowledge can achieve state-of-the-art performance on various few-shot tasks. Nevertheless, human-written label descriptions can be sub-optimal, \citet{Chai20Description} utilize reinforcement learning to generate label descriptions.

\paragraph{Comparison with Prompt-Based Learning}
In both paradigms (\texttt{(M)LM} and \texttt{Matching}), the goal is to reformulate the downstream tasks into the pre-training task (language modeling or entailment). To achieve this, both of them need to modify the input text with some templates to prompt the pre-trained language or entailment model. In prompt-based learning, the prediction is conducted by the pre-trained MLM head on the \texttt{[MASK]} token, while in matching-based learning the prediction is conducted by the pre-trained classifier on the \texttt{[CLS]} token. In prompt-based learning, the output prediction is over the vocabulary, such that a verbalizer is required to map the predicted word in vocabulary into a task label. In contrast, matching-based learning can simply reuse the output (Entailment/Contradiction/Neutral, or Entailment/NotEntailment). Another benefit of matching-based learning is that one can construct pairwise augmented data to perform contrastive learning, achieving further improvement of few-shot performance. However, matching-based learning requires large-scale human annotated entailment data to pre-train an entailment model, and domain difference between the source domain and target domain needs to be handled. Besides, matching-based learning can only be used in understanding tasks while prompt-based learning can also be used for generation~\cite{Li20Prefix,Liu21PTuning}.

\subsection{\texttt{MRC}}
\texttt{MRC} is also an alternative paradigm to unify various NLP tasks by generating task-specific questions and training a MRC model to select the correct span from the input text conditioned on the questions. Take NER as an example, one can recognize the organization entity in the input "\textit{Google was founded in 1998}" by querying a MRC model with "\textit{Google was founded in 1998. Find organizations in the text, including
companies, agencies and institutions}" as in \citet{Li20MRC}. In addition to NER, MRC framework has also demonstrated competitive performance in entity-relation extraction~\cite{Li19Entity}, coreference resolution~\cite{Wu20CorefQA}, entity linking~\cite{Gu2021Read}, dependency parsing~\cite{Gan21Dependency}, dialog state tracking~\cite{Gao2019Dialog}, event extraction~\cite{Du2020Event,Liu2020Event}, aspect-based sentiment analysis~\cite{Mao2021Joint}, \textit{etc}.

\texttt{MRC} paradigm can be applied as long as the task input can be reformulated as \textit{context}, \textit{question}, and \textit{answer}. Due to its universality, \citet{McCann18decaNLP} proposed decaNLP to unify ten NLP tasks including question answering, machine translation, summarization, natural language inference, sentiment analysis, semantic role labeling, relation extraction, goal-oriented dialogue, semantic parsing, and commonsense pronoun resolution in a unified QA format. Different from previously mentioned works, the answer may not appear in the context and question for some tasks of decaNLP such as semantic parsing, therefore the framework is strictly not a \texttt{MRC} paradigm.

\paragraph{Comparison with Prompt-Based Learning}
It is worth noticing that the designed question can be analogous to the prompt in \texttt{(M)LM}. The verbalizer is not necessary in \texttt{MRC} since the answer is a span in the context or question. The predictor, MLM head in the prompt-based learning, can be replaced by a start/end classifier as in traditional MRC model or a pointer network as in \citet{McCann18decaNLP}.

\subsection{\texttt{Seq2Seq}}
\texttt{Seq2Seq} is a general and flexible paradigm that can handle any task whose input and output can be recast as a sequence of tokens. Early work~\cite{McCann18decaNLP} has explored using the \texttt{Seq2Seq} paradigm to simultaneously solve different classes of tasks. Powered by recent advances of seq2seq pre-training such as MASS~\cite{Song2019MASS}, T5~\cite{Raffel20T5}, and BART~\cite{bart}, \texttt{Seq2Seq} paradigm has shown its great potential in unifying diverse NLP tasks. \citet{Paolini2021Structured} use T5~\cite{Raffel20T5} to solve many structured prediction tasks including joint entity and relation extraction, nested NER, relation classification, semantic role labeling, event extraction, coreference resolution, and dialogue state tracking. \citet{Yan2021UniABSA} and \citet{Yan2021UniNER} use BART~\cite{bart}, equipped with the copy network~\cite{Gu2016Copy}, to unify all NER tasks (flat NER, nested NER, discontinuous NER) and all ABSA tasks (AE, OE, ALSC, AOE, AESC, Pair, Triplet), respectively.

\paragraph{Comparison with Other Paradigms}
Compared with other unified paradgms, \texttt{Seq2Seq} is particularly suited for complicated tasks such as structured prediction. Another benefit is that \texttt{Seq2Seq} is also compatible with other paradigms such as \texttt{(M)LM}~\cite{Raffel20T5,bart}, \texttt{MRC}~\cite{McCann18decaNLP}, etc. Nevertheless, what comes with its versatility is the high latency. Currently, most successful seq2seq models are in auto-regressive fashion where each generation step depends on the previously generated tokens. Such sequential nature results in inherent latency at inference time. Therefore, more work is needed to develop efficient seq2seq models through non-autoregressive methods~\cite{Gu18NAR,Qi21BANG}, early exiting~\cite{Elbayad20Depth}, or other alternative techniques.

\section{Conclusion}
\label{sec:conclusion}
Recently, prompt-based tuning, which is to formulate some NLP task into a (M)LM task, has exploded in popularity. They can achieve considerable performance with much less training data. In contrast, other potential unified paradigms, \textit{i.e.} \texttt{Matching}, \texttt{MRC}, and \texttt{Seq2Seq}, are under-explored in the context of pre-training. One of the main reasons is that these paradigms require large-scale annotated data to conduct pre-training, especially \texttt{Seq2Seq} is notorious for data hungry. 

Nevertheless, these paradigms have their advantages over \texttt{(M)LM}: \texttt{Matching} requires less engineering, \texttt{MRC} is more interpretable, \texttt{Seq2Seq} is more flexible to handle complicated tasks. Besides, by combining with self-supervised pre-training (\textit{e.g.} BART~\cite{bart} and T5~\cite{Raffel20T5}), or further pre-training on annotated data with existing language model as initialization (\textit{e.g.} \citet{Wang21Entailment}), these paradigms can achieve competitive performance or even better performance than \texttt{(M)LM}. Therefore, we argue that more attention is needed for the exploration of more powerful entailment, MRC, or seq2seq models through pre-training or other alternative techniques. 

\section*{Acknowledgments}
We would like to thank the members of the FudanNLP group for helpful discussion and valuable feedback. We would also like to thank Wenxuan Zhang from CUHK, and Zengzhi Wang from NUST, for their help in adding important papers. This work was supported by the National Natural Science Foundation of China (No. 62022027).

\bibliographystyle{acl_natbib}
\bibliography{references}
\end{document}